\documentclass[journal]{IEEEtran}

\newcommand{\pp}[1]{\textcolor{black}{#1}}

\usepackage{cite}
\usepackage{xspace}
\usepackage{caption}
\usepackage{subcaption}
\usepackage{xcolor}
\usepackage[cmex10]{amsmath}
\usepackage{mathtools,amssymb,lipsum}
\usepackage{lettrine}
\usepackage{cite}
\usepackage{bm}
\usepackage{booktabs}
\usepackage{algorithm,algorithmic}
\makeatletter
\newcommand{\multilines}[1]{%
	\begin{tabularx}{\dimexpr\linewidth-\ALG@thistlm}[t]{@{}X@{}}
		#1
	\end{tabularx}
}
\makeatother
\usepackage{array}
\usepackage{url}  
\usepackage{amsthm}
\usepackage{multirow}
\usepackage{color}
\usepackage{endnotes}
\usepackage{framed}
\usepackage{cool} % for partial derivative
\usepackage{enumerate} % for \begin{enumerate}[(a)], or [a)]
\usepackage{url}
\usepackage[subnum]{cases}
\usepackage[nocomma]{optidef}
\usepackage{etoolbox}

\usepackage{graphicx}
\DeclareGraphicsExtensions{.pdf,.eps}

\makeatletter
\patchcmd{\@maketitle}
{\addvspace{0.5\baselineskip}\egroup}
{\addvspace{-1.8\baselineskip}\egroup}
{}
{}
\makeatother

\def\BibTeX{{\rm B\kern-.05em{\sc i\kern-.025em b}\kern-.08em
		T\kern-.1667em\lower.7ex\hbox{E}\kern-.125emX}}

\begin{document}

\title{A Contribution-based Device Selection Scheme in Federated Learning}

\author{Shashi Raj Pandey,%
\thanks{This work has been in part supported by the Villum Investigator Grant ``WATER” from the Velux Foundation, Denmark. This work has received funding from the European Union’s Horizon 2020 research and innovation programme under grant agreement No. 957218 (Project IntellIoT). }
\thanks{Shashi Raj Pandey, Lam D. Nguyen, and Petar Popovski are with Connectivity Section, Department of Electronic Systems, Aalborg University, Denmark. Email: \{srp, ndl, petarp\}@es.aau.dk.}
\textit{IEEE Member}, %
 Lam D. Nguyen, \textit{IEEE Member}, %
and Petar Popovski, \textit{IEEE Fellow}}

% The paper headers
\markboth{PREPRINT VERSION}%
{Shell \MakeLowercase{\textit{et al.}}: Bare Demo of IEEEtran.cls for IEEE Journals}

\maketitle
\begin{abstract}
In a Federated Learning (FL) setup, a number of devices contribute to the training of a common model. We present a method for selecting the devices that provide updates in order to achieve improved generalization, fast convergence, and better device-level performance. We formulate a min-max optimization problem and decompose it into a primal-dual setup, where the duality gap is used to quantify the device-level performance. Our strategy combines \emph{exploration} of data freshness through a 
random device selection with \emph{exploitation} through simplified estimates of device contributions. This improves the performance of the trained model both in terms of generalization and personalization. A modified Truncated Monte-Carlo (TMC) method is applied during the exploitation phase to estimate the device's contribution and lower the communication overhead. The experimental results show that the proposed approach has a competitive performance, with lower communication overhead and competitive personalization performance against the baseline schemes.
\end{abstract}

\begin{IEEEkeywords}
federated learning, device selection, exploration, exploitation, personalization, generalization 
\end{IEEEkeywords}

\IEEEpeerreviewmaketitle
\vspace{-0.45cm}
\section{Introduction}
Federated Learning (FL), as introduced by McMahan et. al. \cite{mcmahan2017communication}, discusses unbalanced
and non-i.i.d. (independent and identical distribution) data partitioning across a massive number of unreliable devices, coordinating with a central server, to distributively train learning models without sharing the actual data. In practice, the data samples are generated through device's usage, such as interactions with applications, results to such statistical heterogeneity. Towards that, related works primarily focus on improving the model performance by tackling data properties, i.e., statistical challenges in the FL \cite{mcmahan2017communication, kairouz2021advances}. Noticeably, in the initial work \cite{mcmahan2017communication}, the authors show that their proposed Federated Averaging (FedAvg) algorithm empirically works well with non-i.i.d. data. However, the accuracy of FedAvg varies differently for different datasets, as observed in the existing methods \cite{kairouz2021advances, nishio2019client}, and how client selection is made \cite{nishio2019client,pandey2020crowdsourcing,yoshida2020mab}. For instance, the authors in \cite{nishio2019client, pandey2020crowdsourcing, pandey2021edge} discussed the impact of having heterogeneous clients, given time requirements for per round training execution, during the decentralized model training over unreliable wireless networks. In doing so, more devices are packed within a training round to improve model performance; however, this lead to consumption of excessive communication resources and larger communication rounds to attain a level of global model accuracy. Also, all received local updates are directly aggregated during model aggregation \cite{mcmahan2017communication,pandey2021edge,nishio2019client,xia2020multi}; thus, fairly ignoring their individual contributions and the rationale behind selecting them. In line with that, the authors in \cite{mcmahan2017communication, kairouz2021advances, nishio2019client} revealed that adding local computations can dramatically increase communication efficiency and improve the trained model performance. However, this additional computational load may be prohibitive for some devices.

In principle, the aforementioned issues appear primarily as a result of selecting ill-conditioned devices in the training procedure, without evaluating their marginal contribution in improving the model performance. In fact, in FL literature \cite{kairouz2021advances, nishio2019client,pandey2020crowdsourcing}, device selection problem over wireless networks has remained an overarching challenge, particularly, due to two reasons: \emph{first}, owing to the consequences of statistical and system-level heterogeneity, uniformly random selection of device may lead to slower convergence and poor model performance across devices \cite{pandey2021edge,nguyen2020self}; \emph{second}, devices with trivial contributions may get scheduled in model training process which only adds larger communication overhead \cite{kairouz2021advances}. Moreover, device's contribution are unknown a priori, and its estimation is not straightforward. Some recent works \cite{yoshida2020mab, xia2020multi} introduced quality-aware device selection strategy with multi-arm bandits (MAB) method in the absence of estimates on available computation-communication resources with fixed dataset; however, they focus on minimization of the convergence time, leaving aside the impact of device selection on the trained model performance, particularly on unseen data, as well as the device-level performance.  

\begin{figure}[t!]
    \centering
    \includegraphics[width=\linewidth]{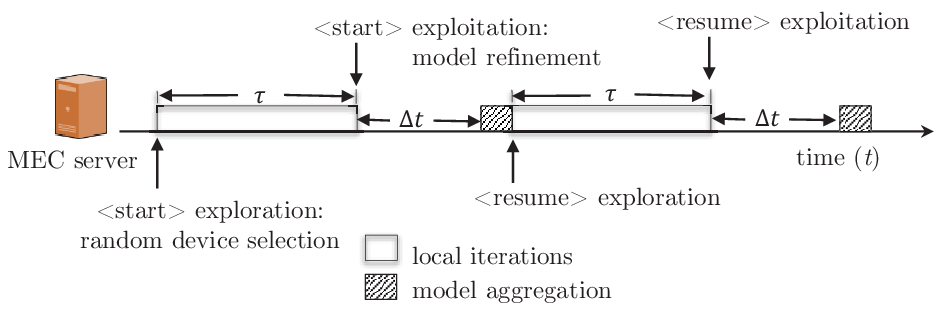}
    \caption{An illustration of device selection strategy with a mix of exploration and exploitation.}
    \label{fig:sys_model}
\end{figure}
In this work, we are interested in, and provide a solution to, the problem of device selection and its impact on model performance defined in terms of \emph{generalization}, i.e., how the trained model work on unseen/new data samples, and \emph{personalization}, i.e., the device-level performance. Different from the related literature, our approach considers the impact of device selection procedure in the decentralized training framework of FL with fairness guarantees in the model performance. An example, compared with \cite{yang2020age,guo2021dynamic}, where fairness is associated with device scheduling opportunities following statistical and/or system-level heterogeneity, our perspective and definition on fairness guarantees is unique. The term \emph{fairness} captures the discrepancies in device-level model performance, i.e., personalization performance, appeared as a direct consequence of statistical heterogeneity and device selection approach during model aggregation. Referring to Fig.~\ref{fig:sys_model}, similar to FedAvg \cite{mcmahan2017communication}, we start by random device selection in FL during the exploration phase at each global iteration. Subsequently, in the exploitation phase, the focus is on contribution-based model aggregation for refining the global model, which, although fundamental, is not addressed by FedAvg. As the contributions of devices in enhancing the global model performance is unknown a priori, we develop a mechanism that first provides an estimation of device's contribution, which is followed by model aggregation with the objective of ensuring fairness across selected device. This way, we bring balance in improving both generalization and device-level performance of the trained model. The summary of our contributions are as follows.

\begin{itemize}
\item We develop a cost-efficient, simplified device selection mechanism in the FL setting that captures the notion of fairness in terms of personalized performance. It lowers the communication overhead while enhancing the generalization performance of the trained model. Therein, as an initial step, the problem is revealed as a min-max optimization problem that  specifically captures the aforementioned joint objectives.
\item To solve the problem of device selection, we propose a strategic mix of exploration and exploitation during each global iteration. This is in contrast to periodic exploration that particularly leads to poor model performance. In particular, we unleash a mechanism that both assess and incorporates the value of contribution of each randomly selected device during model training.
\item We propose a contribution-based device selection (CDS) technique for model aggregation with the modified Truncated Monte-Carlo (TMC) \cite{jia2019towards} method during exploitation phase within random device selection rounds, i.e., the exploration phase. We show this technique eliminates the need of adding local computations to lower communication overhead and improve model convergence.
\item The simulation results show that the proposed method, while offering a $4$X decrease in the communication rounds, still brings a small performance improvement in terms of personalization.
\end{itemize}

Overall, the chief contribution of this work is that it 
incorporates a contribution-based device selection strategy for model aggregation in FL. 
This improves device-level performance in several aspects simultaneously:  generalization, low communication overhead, and fast convergence.

\vspace{-0.45cm}
\section{Problem Setting}
We consider a supervised learning problem in a FL setting where each available device (client) $m \in \mathcal{M}$ in the training process has a local data set $\mathcal{D}_m $ of size $D_m$ such that the overall training data size $D=\sum_{m=1}^{M}D_{m}$. The devices communicate with the central aggregator (e.g., a multi-access edge computing (MEC) server) via orthogonal frequency division multiple access (OFDMA). For simplicity, downlink is treated as error-free. The data samples are defined as a collection set of input-output pairs $\{x_i, y_i\}_{i=1}^{D_m} $, where, respectively, $x_i \in \mathcal{X}$ characterizes the feature space of data sample with corresponding labels $y_i \in \mathcal{Y}$, $\mathcal{X} \subseteq \mathbb{R}^d$ and $\mathcal{Y} \subseteq \mathbb{R}$. \pp{The goal of the} central aggregator orchestrating the distributed training process is to learn a predictor $F (w):\mathcal{X} \rightarrow \mathcal{Y}\in \mathbb{R}$. The predictor performance is associated with learning of a model parameter $w \in \mathbb{R}^d$ that captures the empirical risk on all distributed data samples as the regularized finite-sum objective of the form 
\begin{equation}
\underset{w \in \mathbb{R}^d}{\text{min}}J(w)+\lambda g(w) \ \  \text{where} \ \ J(w) := \sum\nolimits_{m = 1}^M \frac{D_m}{D}\cdot J_m (w),
\label{eq:learning_problem}
\end{equation}
$J_{m}(w) := \frac{1}{D_m} \sum \nolimits_{i = 1}^{D_m}f_i(w)$, $f_i(w)$ is the loss function that depends upon the choice of learning algorithms \cite{mcmahan2017communication, pandey2021edge}; $\lambda$ is a regularizer and $g(w)$ is a well-known regularization function that controls the complexity and captures the device-level performance. We assume $f_i$ is $(1 / \gamma)$-smooth and 1-strongly convex of $g$ and continuous differentiable to ensure convergence of the solution; a common assumption made in several existing works \cite{hanzely2020lower, pandey2021edge,boyd2004convex}.

\textbf{Solving FL problem with FedAvg \cite{mcmahan2017communication}:}
The learning problem in \eqref{eq:learning_problem} is solved following a two-step approach in an iterative manner. \textit{Step 1}: In the $t^{\textrm{th}}$ update of the global iteration, a fraction of selected devices $m$ iterates over its on-device training data $\mathcal{D}_m$ to solve the local problem 
	\begin{equation}
		w^{(t)}_m = \underset{w_m \in \mathbb{R}^{d}}{\mathrm{arg\, min}} \ J_m (w_m | w^{(t-1)}),
	\end{equation}
and communicates the parameters $w^{(t)}_m$ to MEC. 
\textit{Step 2}:At the MEC, the collected local parameters are simply aggregated as 
	\begin{align}
	& w^{(t+1)} = \sum\nolimits_{m = 1}^M \frac{D_m}{D}\cdot w^{(t)}_m, \label{eq:global_weight}
	\end{align}
	and broadcast back to the fraction of participating devices for the next global iteration.
This process continues until a desired level of model accuracy is achieved. We observe, this approach of model aggregation imposes an equal value of contribution to the selected local model parameters without considering its consequences on the model performance; e.g. a deterioration due to poor local updates. This eventually exerts larger communication rounds to converge to a high-quality global model while alongside inducing irregularities in the device-level performances.
\vspace{-0.5cm}
\subsection{Device Selection in FL with Fairness Guarantees}
We note that a plain FL \cite{mcmahan2017communication} ignores the device-level performance, particularly caused due to small data, limited computational capabilities and available communication channels to share parameters with the server, or having devices with poor local updates. Hence, the random device selection procedure may incur costs in terms of overhead. While it is also significant to maintain device-level performance, the cost overhead due to uncertainty in random device selection should be minimized. Therein, we first develop a natural extension to plain FL which captures the risk associated with the local accuracy loss using model $w$ in each device $m$ as
	\begin{equation}
		R^{(t)}_m(F)= \underset{\{X^m,Y^m\}\sim \mathcal{P}_m}{\mathbb{E}} \Bigg[J(F(X^m; w), Y^m) \Bigg],
	\end{equation}
where $\mathcal{P}_m$ is the underlying distribution of local data samples. Here, we look forward to characterize a predictor $F$ that works well with dataset in all associated devices. Basically, in a risk-averse FL setting, we aim to minimize the maximum risk across all participating devices, ensuring improved personalization and generalization performance of the trained model. Subsequently, we formulate a min-max optimization problem: it jointly selects the devices contributing the most in improving the model performance and minimize the maximal risk to ensure fairness amongst these associated devices: 
	\begin{align}
	{\textbf{P:}} \min_{w} \quad 
	& \Bigg[\max_{\tilde{\mathcal{M}}:=\{m\}} \sum\nolimits_{m \in \tilde{\mathcal{M}}} \ R^{(t)}_m(F)  \Bigg],  \\
	\text{s.t.}\quad & w \in  \arg \min_{w \in \mathbb{R}^{d}} R_m^{(t)}(F_m) , \quad \forall m \in \tilde{\mathcal{M}}, \\
	& \quad R_m^{(t)}(F_m) \le \theta_m^{\textrm{th}}, \forall m \in \tilde{\mathcal{M}} \label{cons:fairness},\\
	& \quad \tilde{\mathcal{M}} \subseteq \mathcal{M}.
	%		\end{alignat}
	\end{align}
where $\tilde{\mathcal{M}}$ is a subset of devices selected based on their local contribution to the global model and satisfying fairness constraint  \eqref{cons:fairness} imposed as $\theta_m^{\textrm{th}}$, defined with details in Section \ref{ssec:device contribution}. To solve \textbf{P}, we need to efficiently derive $\tilde{\mathcal{M}}$ in prior, which is not straightforward. For that purpose, we use the primal-dual decomposition method and capture the contribution of devices in terms of the quality of obtained dual solution, that is defined in details in the following Section~\ref{ssec:device contribution}. 

Next, we define and discuss the incurred time costs from the perspective of random device selection procedure.
\vspace{-0.5cm}
\subsection{Time cost} 
In each global iteration, the time cost for device $m$ has two components: (i) local computation time: $t_m^{\textrm{comp}} = c_mD_m/f_m$, where $c_m$ is CPU-cycles requirement to compute per bit data; $f_m$ is CPU-cycle frequency; and (ii) communication time: $t_m^{\textrm{comm}} = s_m/(r_mB_m)$, where $s_m$ is the size of transmitted data in bits; $r_m = \log_2(1+ \frac{p_mh_m}{N_0B_m})$ is the rate; $p_m$ is the transmission power; $h_m$ is the channel gain; $N_0$ is the noise power spectral density and $B_m$ is the corresponding bandwidth. Then, the corresponding execution time cost $t^{\textrm{exec}}$ per global iteration due to random device scheduling is proportional to $t^{\textrm{cost}}:=\underset{m}{\max}\{t_m^{\textrm{comp}}+t_m^{\textrm{comm}}\}, \forall m \in \mathcal{M}.$ This \pp{captures the fact that minimization of $t_\tau^{\textrm{cost}}$ while ensuring fairness of the collaborative training in solving \textbf{P}}, there is a need for careful device selection $\tilde{\mathcal{M}}$ and subsequent model aggregation. 

In the following, we first reflect the risk minimization problem \textbf{P} for a known set of selected devices in a primal-dual setting. Then, we propose a cost-efficient contribution-based device selection approach with fairness guarantees by exploiting the modified TMC \cite{jia2019towards, nguyen2021marketplace} method to solve the \textbf{P} in subsequent global iterations.

\vspace{-0.45cm}
	\begin{algorithm}[t!]
        	\caption{\strut CDS: Contribution-based Device Selection}
        	\begin{algorithmic}[1]
        	   \STATE{\texttt{Input}: $M$ is the number of devices in set $\mathcal{M}$; $C \in (0,1]$ is fraction of random devices selected; $B$ is the local minibatch size; $E$ is the number of epochs, and $\eta$ is the learning rate.}
        	   \STATE{\texttt{Initialize:} $\phi^1, C, B, E, \Delta t, \epsilon$ and $\eta$.}\\
        	   \texttt{//Exploration://}\\
        	   \FORALL{global iteration $t \in \{1,2,\ldots, \tau\}$}
        	   \STATE{Set $\mathcal{S}_t  \subseteq \mathcal{M}$ with $\max(1, C\times M)$ devices};
            	   \FORALL{$m \in \mathcal{S}_t$}
            	   \STATE{Execute \texttt{DeviceUpdate}($m,\phi^t$)};
            	   \ENDFOR\\
               \texttt{//Exploitation://}
               \STATE Start exploitation timer $t'=0$;
               \WHILE {$t'<\Delta t$}
               \STATE Define $\pi^{t'}$ as random permutation of the selected devices;
               \STATE Set $v^{t'}_0 = V(\emptyset)$;
               \FORALL{$m \in \mathcal{S}_{t'}$}
                \IF{$|V(\phi^{t'}) - v^{t'}_{m-1}|< \epsilon$}
                    \STATE $v^{t'}_m = v^{t'}_{m-1}$;
                    \ELSE
                    \STATE Set $v^{t'}_m \leftarrow V(\{\pi^{t'}[1], \ldots,\pi^{t'}[m]\});$
                \ENDIF
                \STATE $\beta_{\pi^{t'}[m]} \leftarrow \frac{{t'}-1}{{t'}}\beta_{\pi^{{t'}-1}[m]} + \frac{1}{{t'}}(v^{t'}_m-v^{t'}_{m-1})$;
               \ENDFOR
               \STATE Update $t'= t'+1$;
               \ENDWHILE
               \STATE Sort $\beta_{\pi^{t'}[m]}$ in descending order; 
               \STATE Obtain $\tilde{\mathcal{M}}$ as per contributions $\beta_{\pi^{t'}[m]}$ and update global variable $\phi^t$ \eqref{eq:global_variable};
        	   \ENDFOR \\
        	   \texttt{DeviceUpdate}($m,\phi^t$): 
        	   \STATE Solve the local sub-problem \eqref{eq:learning_problem_local};
        	   \STATE Update dual variables using \eqref{eq:local_variable};
        	   \STATE \textbf{Return} $\Delta \phi^t_{[m]}, \forall m \in \mathcal{M};$
        	\end{algorithmic}
        	 \label{alg:algo}
        \end{algorithm}

\section{Contribution-based Device Scheduling Algorithm}\label{sec:contribution based scheduling}
\subsection{Device's local contribution in a distributed setting}\label{ssec:device contribution}
We revisit the global problem \eqref{eq:learning_problem} in its dual optimization form \cite{boyd2004convex} with $M$ devices to evaluate their local contribution in solving the learning problem. The corresponding dual optimization problem of \eqref{eq:learning_problem} for a convex loss function $f$ is 
	\begin{equation}
	\underset{\alpha \in \mathbb{R}^{D}}{\text{max}} \boldsymbol{\mathcal{R}}(\alpha) := \frac{1}{D} \sum \nolimits_{i = 1}^{D} - f_i^*(- \alpha_i) - \lambda g^*(\phi(\alpha)),
	\label{eq:learning_problem_dual}
	\end{equation}	
	where $f_i^*$ and $g^*$ are the convex conjugates of $f_i$ and $g$, respectively \cite{boyd2004convex}; $\alpha \in \mathbb{R}^{D}$ is the dual variable mapping to the primal candidate vector; and $\phi(\alpha) = \frac{1}{\lambda D} X \alpha$. Here, we define $X \in \mathbb{R}^{d \times D_m}$ as a matrix with columns having data points for $i \in \mathcal{D}_m, \forall m$. Then, having the optimal value of dual variable $\alpha^*$ in \eqref{eq:learning_problem_dual}, we obtain the optimal solution\footnote{Finding the optimal solution follows an iterative process to attain a global accuracy $0 \le \epsilon \le 1$ (i.e., $\mathbb{E}\left[ \boldsymbol{\mathcal{R}}(\alpha) - \boldsymbol{\mathcal{R}}(\alpha^*)\right] < \epsilon$).} of \eqref{eq:learning_problem} as $w(\alpha^*) = \nabla g^*(\phi(\alpha^*))$ \cite{pandey2020crowdsourcing}. Hereafter, we use $\phi \in \mathbb{R}^d$ for $\phi{(\alpha)}$ for simplicity, and define a weight vector $\varrho_{[m]} \in \mathbb{R}^D$ at the local sub-problem $m$ with its elements zero for the unavailable data points.
	As the consequence of the properties of $f_i$ and $g$, we obtain the approximate solution to the local sub-problem:
	\begin{equation}
	\underset{\varrho_{[m]}\in \mathbb{R}^{D}}{\text{max}} \boldsymbol{\mathcal{R}}_m (\varrho_{[m]}; \phi, \alpha_{[m]}),
	\label{eq:learning_problem_local}
	\end{equation}
	defined by the dual variables $\alpha_{[m]}$, $\varrho_{[m]}$. Here, we have $\boldsymbol{\mathcal{R}}_m (\varrho_{[m]}; \phi, \alpha_{[m]}) = -\frac{1}{M} - \langle \nabla (\lambda g^*(\phi(\alpha))), \varrho_{[m]} \rangle  - \frac{\lambda}{2} \Vert\frac{1}{\lambda D}X_{[m]}\varrho_{[m]}  \Vert^2$ with $X_{[m]}$ as a matrix with columns having data points for $i \in \mathcal{D}_m$, and zero padded otherwise.
	Each selected device $m \in \mathcal{M}$ iterates over its computational resources using any arbitrary solver to solve its local problem \eqref{eq:learning_problem_local} for a defined local relative $\theta_m^{\textrm{th}}$ accuracy, i.e., the maximal risk in \textbf{P} in terms of fairness, that characterizes the quality of the local solution and produces a random output $\varrho_{[m]}$ satisfying
	\begin{equation}
	\mathbb{E}\left[ \boldsymbol{\mathcal{R}}_m(\varrho^*_{[m]}) -  \boldsymbol{\mathcal{R}}_m(\varrho_{[m]})  \right] \le \theta_m^{\textrm{th}} \left[  \boldsymbol{\mathcal{R}}_m(\varrho^*_{[m]}) -  \boldsymbol{\mathcal{R}}_m(0) \right]. 
	\end{equation}
	Then, the local dual variable is updated as follows:
	\begin{equation}
	\alpha^{t+1}_{[m]} := \alpha^t_{[m]} + \varrho^t_{[m]},\forall m \in \mathcal{M}.
	\label{eq:local_variable} 
	\end{equation}
	The selected devices then broadcast the local parameter defined as $\Delta \phi^t_{[m]}:= \frac{1}{\lambda D} X_{[m]}\varrho^t_{[m]} $ to the MEC server along with its local relative accuracy, which completes the exploration phase. Next, considering all selected devices contribute equally, which is the usual case of FedAvg, the MEC server can simply aggregate the local parameters as
	\begin{equation}
	\phi^{t+1} := \phi^t + \frac{1}{M}\sum \nolimits_{m = 1}^{M} \Delta \phi^t_{[m]}.
	\label{eq:global_variable}
	\end{equation}
	and share $\phi^{t+1}$ back to the randomly selected devices to again solve \eqref{eq:learning_problem_local}. However, instantiating random device selection in each exploration phase with the global model using FedAvg demerits local model performance, particularly poor personalization and large variance in model accuracy across devices. This is because the technique does not specifically consider the actual contribution of each device in the model training, and tune parameters accordingly to minimize the empirical risk at worse performing devices; \pp{this leads to the increase of the number of} communication rounds for convergence. Therefore, we depart from the naive approach of model aggregation and introduce an additional exploitation phase, where the contributions made by selected devices \pp{are} taken into consideration. 
	
	In the following, we present our proposed approach that utilizes the contribution of each device within the exploration phase in improving the model performance and lowering the long-term communication overhead.
	\vspace{-0.5cm}
\subsection{Contribution-based device selection with modified TMC}
Within each round of global iteration that executes random device scheduling, we first eliminate the frequency of transmitting the local dual variables through an estimation of TMC at the MEC, leading to an efficient parameter aggregation scheme. Specifically, following the value of contribution of each selected device, we operate $\Delta t$ rounds of random permutations on received dual variables to perform the model aggregation before proceeding to the next global iteration with random device selection strategy. We use the estimation of the contribution of each selected device $m \in \mathcal{S}_t$ in round $t$ as
\begin{equation}
    \beta_m = \frac{1}{|\mathcal{S}_t|!}\sum\nolimits_{\pi \in \Pi(\Delta\phi^t_{[m]})}[V(P^{\pi}_m \cup \{m\}) - V(P_m^{\pi})],
    \label{eq:shapTMC}
\end{equation}
where $V(\cdot)$ is the standard valuation of local parameter defined as its performance score while contributing to improve the model accuracy, similar to \cite{jia2019towards}. $\pi \in \Pi(\Delta\phi^t_{[m]})$ characterizes the permutation of devices and $P^{\pi}_m$ is the preceding devices selected for model aggregation. The details are described in Algorithm~\ref{alg:algo}. Once the exploration phase is executed, using \eqref{eq:shapTMC}, the MEC proceeds to the exploitation phase (lines 8--21) to estimate the contributions of each selected device parameters in improving the model performance (lines 12--20). Then, the devices are sorted based on their contributions, after which the final model is obtained (line 23) to execute the next round of exploration.

\vspace{-0.45cm}
\begin{figure*}[t!]
     \centering
     \begin{subfigure}[b]{0.24\textwidth}
         \centering
         \includegraphics[width=0.8\linewidth]{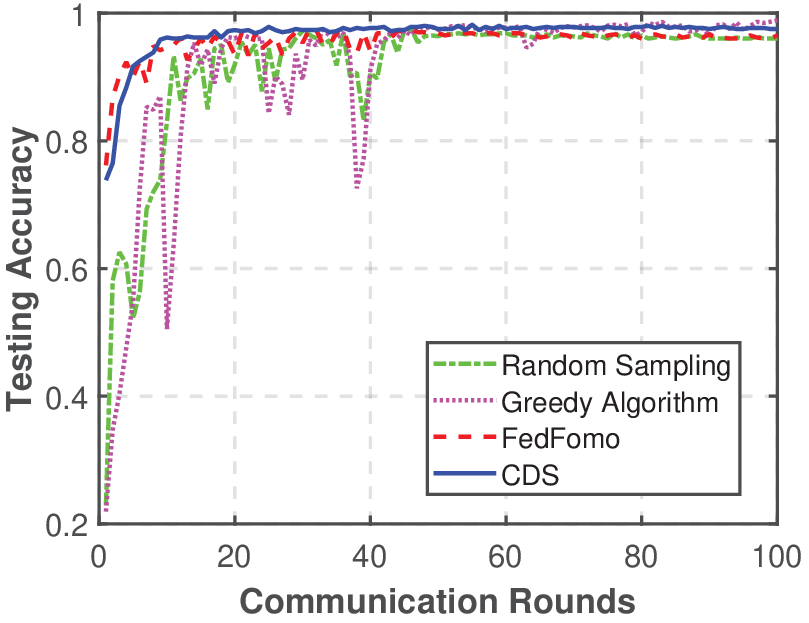}
         \caption{Testing accuracy.}
         \label{fig:2a}
     \end{subfigure}
     \begin{subfigure}[b]{0.24\textwidth}
         \centering
         \includegraphics[width=0.8\linewidth]{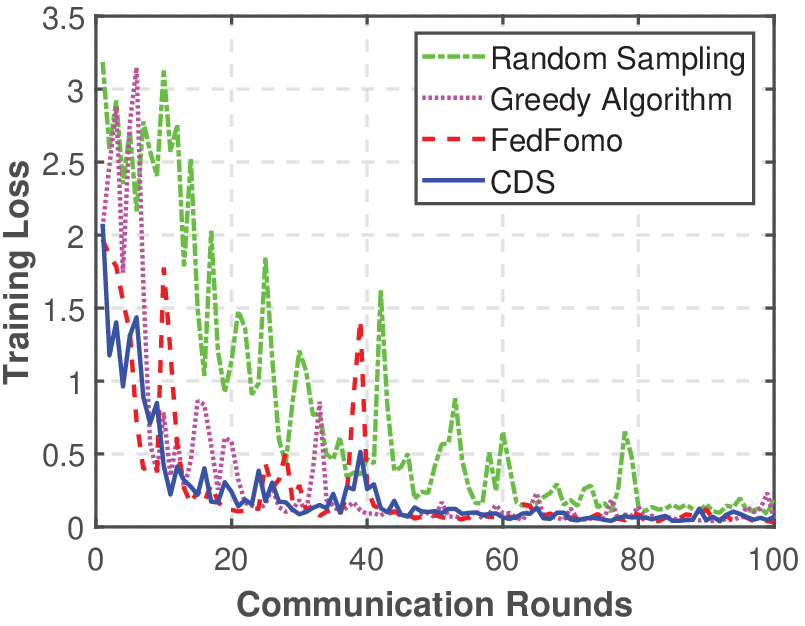}
         \caption{Training loss.}
         \label{fig:2b}
     \end{subfigure}
     \begin{subfigure}[b]{0.24\textwidth}
         \centering
         \includegraphics[width=0.8\linewidth]{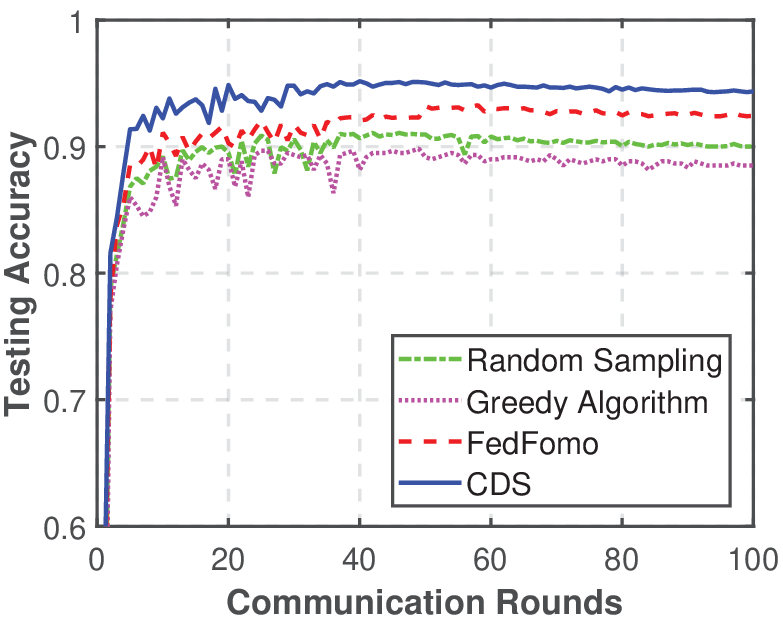}
         \caption{Device-level performance.}
         \label{fig:2c}
     \end{subfigure}
          \begin{subfigure}[b]{0.24\textwidth}
         \centering
         \includegraphics[width=0.8\linewidth]{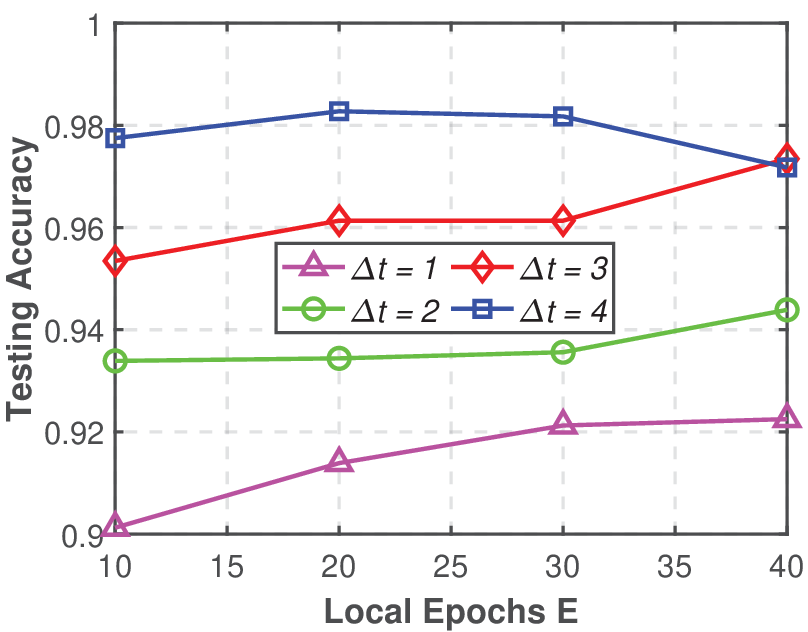}
         \caption{Effect of $\Delta t$.}
         \label{fig:2d}
     \end{subfigure}
     \caption{Comparative performance analysis with different sampling strategies on: (a) testing accuracy, (b) training loss, (c) personalization, i.e., the device-level performance, and (d) the impact of $\Delta t$ against adding more local computations during local training.  }
     \label{fig:simu}
\end{figure*}

\section{Performance Evaluation}
We have simulated the MEC environment with network parameters defined in \cite{nishio2019client} and conducted extensive experiments with well-known MNIST datasets \cite{MNISThan75:online}. We have taken into account the statistical heterogeneity involving the non-i.i.d. characteristics and unbalanced dataset, in order to appropriately setup data across the devices \cite{pandey2021edge}. We shuffle and divide dataset at each device, where data samples are drawn particularly from two labels out of ten for MNIST. We have considered $M=100$ devices on a single server with 26 core Intel Xeon 2.6 GHz, 256 GB RAM, 4 TB, Nvidia V100 GPU, Ubuntu OS. Accordingly, we have normalized the time cost per global iteration and also assumed that the devices and server are synchronized. We have then evaluated the model performance by considering the impact of device selection strategy during the model aggregation. For training, each device uses $E=10$ epochs, the batch size $B=10$, and the learning rate $\eta =0.01$ as a default. Further, in all of our evaluations, unless specified, we set the value of $C=0.1$, which exhibits a good balance between computational efficiency and convergence rate, as demonstrated experimentally in \cite{mcmahan2017communication}. The default value of $\Delta t$ is set to 1. We compare CDS against (a) two intuitive baselines: 1) Random Sampling method which choose devices randomly, 2) Greedy Algorithm \cite{balakrishnan2021diverse}, in which the server collects the local gradients of all devices and chooses a subset of devices providing the greatest marginal gain, and (b) a more competitive approach, namely FedFomo\cite{zhang2020personalized}, customized for personalized federated updates of each device.  

Fig.~\ref{fig:2a} shows that that our proposed contribution-based device selection approach achieves a high accuracy, approximately 97\%, while using substantially fewer communication rounds compared to the Greedy and Random Sampling algorithms. This is due to the fact that we are strategically selecting device updates based on their contribution in improving the model performance during model aggregation. This is further supported by Table~\ref{tab:exec cost}, where we present a comparative analysis on the execution time cost $t^{\textrm{exec}}$ in achieving around $80\%$ of the target device-level model accuracy and the required number of communication rounds. We note a relaxed threshold accuracy allows a larger set of participating devices, and thus, leads better generalization during model training. In all cases, we see CDS demonstrates better and stable performance than these baselines, and a competitive model accuracy as compared with FedFomo. Similar observations pertain to the training loss of different sampling strategies, as shown in Fig.~\ref{fig:2b}. In Fig.~\ref{fig:2c}, we evaluate and compare the average device-level performance as the performance of the trained model on their local test dataset. Interestingly, we see that CDS provides improvement in the device-level performance: $6.54\%$ and $4.78\%$ higher than Greedy method and Random sampling, respectively, as well as $2.54\%$ compared to FedFomo. This is due to the appropriateness of the local updates that are aggregate in the model following the contributions of the selected devices in every global round. Furthermore, Random Sampling shows better performance in terms of personalization compared to the Greedy. This is reasonable as it explores the available devices to improve personalization, while the Greedy approach focuses only on improving the performance of the global model.

Finally, in Fig.~\ref{fig:2d}, we analyze the significance of $\Delta t$ on eliminating the time cost required for the model convergence to a target accuracy. The insight is that, instead of adding more computation per device to lower the communication overhead \cite{mcmahan2017communication}, we perform contribution-based model aggregation that compensates the increase in local computation costs and the resulting higher $t^\textrm{cost}$ for achieving better convergence. Specifically, relaxing $\Delta t$ allows more rounds of random permutation to explore the contribution of local model updates. In return, having broader exposure of the contributing devices results in better device selection strategy for model aggregation and a better model performance as shown in Fig.~\ref{fig:2d}, while saving local computations (epoch) at the devices. We also observe the impact of relaxing $\Delta t$ as a reference to evaluate and compare model performance against adding more local computations. For example, when $E=40$, we see no significant improvement in the model performance even though we increase $\Delta t$ by more than $3$ rounds, which is due to the optimized device selection strategy.  
\begin{table}[t!]
\centering
\caption{Comparison of execution time cost $t^{\textrm{exec}}$ (in $sec$).}
\begin{tabular}{ccc}
\toprule
\multirow{2}{*}{\textbf{Algorithms}} & \multicolumn{2}{l}{\textbf{\begin{tabular} [c]{@{}l@{}}Average time to achieve accuracy of \\ 80\%, and number rounds to achieve\end{tabular}}} \\ \cline{2-3} 
                                     & \multicolumn{1}{l}{ Average $t^{\textrm{exec}}$}                   & \multicolumn{1}{l}{Number of rounds}                                \\ \midrule 
Random Sampling \cite{mcmahan2017communication}           & 0.521664                      & 5$\pm$1                                     \\
Greedy-based solution\cite{balakrishnan2021diverse}       & 1.768358                      & 9$\pm$1                                          \\
FedFomo\cite{zhang2020personalized}                      & 0.338333                       & 2$\pm$1                                    \\
CDS - Our proposed                                       & 0.335492                       & 2$\pm$1                              \\

\bottomrule
\end{tabular}
\label{tab:exec cost}
\end{table}
\vspace{-0.45cm}
\section{Conclusion}
We have presented a simplified solution of the device selection problem in Federated Learning (FL), aiming to jointly improve model performance and lower the communication costs. In doing so, we have first formulated a min-max optimization problem. to solve it, we have developed a strategy that constitutes a mixture of exploration phase, where random selection of devices is made, similar to the plain FL approach but under a primal-dual setting of the learning problem, and an additional exploitation scheme that quantifies the contribution of selected devices in improving the model performance via efficient model aggregation. Extensive simulations on real-world dataset have demonstrated the efficacy of the proposed approach against the baselines in improving model performance, i.e., better generalization and personalization, lowering communication costs, and achieving fast convergence rate.
\vspace{-0.5cm}
\bibliographystyle{ieeetr}
\bibliography{ref}

\begin{thebibliography}{10}

\bibitem{mcmahan2017communication}
B.~McMahan {\em et~al.}, ``Communication-efficient learning of deep networks
  from decentralized data,'' in {\em International Conference on Artificial
  Intelligence and Statistics (AISTATS)}, April 2017.

\bibitem{kairouz2021advances}
P.~Kairouz {\em et~al.}, ``Advances and open problems in federated learning,''
  {\em Foundations and Trends in Machine Learning}, vol.~14, no.~1--2,
  pp.~1--210, 2021.

\bibitem{nishio2019client}
T.~Nishio and R.~Yonetani, ``Client selection for federated learning with
  heterogeneous resources in mobile edge,'' in {\em IEEE International
  Conference on Communications (ICC)}, May 2019.

\bibitem{pandey2020crowdsourcing}
S.~R. Pandey {\em et~al.}, ``A crowdsourcing framework for on-device federated
  learning,'' {\em IEEE Transactions on Wireless Communications}, vol.~19,
  no.~5, pp.~3241--3256, Feb. 2020.

\bibitem{yoshida2020mab}
N.~Yoshida {\em et~al.}, ``Mab-based client selection for federated learning
  with uncertain resources in mobile networks,'' in {\em IEEE Globecom
  Workshops (GC Wkshps)}, Dec. 2020.

\bibitem{pandey2021edge}
S.~R. Pandey {\em et~al.}, ``Edge-assisted democratized learning toward
  federated analytics,'' {\em IEEE Internet of Things Journal}, vol.~9, no.~1,
  pp.~572--588, June 2021.

\bibitem{xia2020multi}
W.~Xia {\em et~al.}, ``Multi-armed bandit-based client scheduling for federated
  learning,'' {\em IEEE Transactions on Wireless Communications}, vol.~19,
  no.~11, pp.~7108--7123, July 2020.

\bibitem{nguyen2020self}
M.~N. Nguyen {\em et~al.}, ``Self-organizing democratized learning: Towards
  large-scale distributed learning systems,'' {\em arXiv preprint
  arXiv:2007.03278}, 2020.

\bibitem{yang2020age}
H.~H. Yang {\em et~al.}, ``Age-based scheduling policy for federated learning
  in mobile edge networks,'' in {\em IEEE International Conference on
  Acoustics, Speech and Signal Processing (ICASSP)}, 2020.

\bibitem{guo2021dynamic}
K.~Guo {\em et~al.}, ``Dynamic scheduling for heterogeneous federated learning
  in private 5g edge networks,'' {\em IEEE Journal of Selected Topics in Signal
  Processing}, 2021.

\bibitem{jia2019towards}
R.~Jia {\em et~al.}, ``Towards efficient data valuation based on the shapley
  value,'' in {\em International Conference on Artificial Intelligence and
  Statistics (AISTATS)}, April 2019.

\bibitem{hanzely2020lower}
F.~Hanzely {\em et~al.}, ``Lower bounds and optimal algorithms for personalized
  federated learning,'' {\em Advances in Neural Information Processing
  Systems}, vol.~33, pp.~2304--2315, 2020.

\bibitem{boyd2004convex}
S.~Boyd {\em et~al.}, {\em Convex optimization}.
\newblock Cambridge university press, March 2004.

\bibitem{nguyen2021marketplace}
L.~D. Nguyen {\em et~al.}, ``A marketplace for trading ai models based on
  blockchain and incentives for iot data,'' {\em arXiv preprint
  arXiv:2112.02870}, 2021.

\bibitem{MNISThan75:online}
``Mnist handwritten digit database, yann lecun, corinna cortes and chris
  burges.'' \url{http://yann.lecun.com/exdb/mnist/}.
\newblock (Accessed on 03/02/2022).

\bibitem{balakrishnan2021diverse}
R.~Balakrishnan {\em et~al.}, ``Diverse client selection for federated
  learning: Submodularity and convergence analysis,'' in {\em International
  Conference on Machine Learning Society (ICML)}, July 2021.

\bibitem{zhang2020personalized}
M.~Zhang {\em et~al.}, ``Personalized federated learning with first order model
  optimization,'' in {\em International Conference on Learning Representations
  (ICLR)}, May 2021.

\end{thebibliography}
\end{document}